\documentclass{article}
\usepackage{spconf,amsmath,amssymb,graphicx,booktabs,array,multirow,url}
\usepackage[hidelinks]{hyperref}

\newcommand{\cmark}{\ensuremath{\checkmark}}
\newcommand{\xmark}{\ensuremath{\times}}
\title{ReDiffNet: Differential RGB-Infrared Learning for Low-Light UAV Oriented Vehicle Detection}

\name{
Qifan Zhang$^{1}$,
Ziran Zhou$^{1}$,
Ruijie Li$^{2}$,
Jincheng Tang$^{3}$,
Hao Wang$^{1}$,
Qihao Qiao$^{1}$,
Chunliu Wang$^{1}$\sthanks{Corresponding author.}
}

\address{
$^{1}$Dalian Maritime University, Dalian, China\\
$^{2}$The Hong Kong University of Science and Technology (Guangzhou), Guangzhou, China\\
$^{3}$Hubei University of Economics, Wuhan, China
}

\begin{document}
\raggedbottom
\ninept
\maketitle
\begin{abstract}

Low-light UAV-based RGB-infrared oriented small-vehicle detection is important for nighttime traffic monitoring, emergency response, and urban inspection. Illumination variations, headlight glare, local shadows, and thermal-response degradation cause spatially varying modality reliability, while the small visual extent of vehicles further weakens boundaries, orientation cues, and thermal responses. Accordingly, selecting trustworthy observations based on local modality reliability while further exploiting complementary discriminative information in regions with ambiguous modality preference is key to constructing effective multimodal representations. Based on this insight, we propose ReDiffNet, a reliability-conditioned differential representation network in which modality reliability guides both evidence selection and complementary recovery. Specifically, degradation-aware reliability learning estimates relative spatial reliability, uncertainty-guided differential recovery exploits cross-modal differences to recover complementary cues in ambiguous regions, and reliability-conditioned reconstruction integrates retained and recovered evidence into a unified representation. ReDiffNet achieves 85.3\% and 73.9\% $\mathrm{mAP}_{50}$ on DroneVehicle and VEDAI, respectively, supporting its effectiveness.
\end{abstract}


\section{Introduction}
Reliable low-light UAV vehicle detection is essential for nighttime traffic monitoring, emergency response, and urban inspection. In this setting, a detector must localize small vehicles with arbitrary orientations from a top-down view, where targets occupy few pixels and their boundaries are easily confused with road markings, shadows, and background clutter~\cite{shakhatreh2019unmanned,tang2023survey,suo2023hit}. RGB-infrared (RGB-IR) imaging is attractive because RGB supplies texture and geometric boundaries, whereas infrared preserves thermal contrast when visible illumination is weak~\cite{r4,r5,r6}.

The central challenge lies in determining which modality provides more reliable evidence at each spatial location. Headlight glare, underexposure, cast shadows, weak thermal contrast, and imperfect local registration affect RGB and infrared differently~\cite{r4,r5,r19}. Thus, one modality may be more informative at a vehicle boundary while the other provides stronger evidence in a dark interior region. Existing RGB-IR detectors mainly emphasize feature alignment, attention-based interaction, or fixed fusion after feature extraction~\cite{r4,r5,r6,r7}. These advances motivate an explicit spatial reliability mechanism that estimates relative modality preference and guides subsequent complementary cue recovery.

We address this issue by casting RGB-IR fusion as reliability-conditioned evidence reconstruction. The proposed ReDiffNet first estimates a spatial field of relative modality reliability using Degradation-Aware Reliability Learning (DARL). It then converts similarity between the two reliability responses into a reliability-ambiguity cue that modulates Uncertainty-Guided Differential Recovery (UGDR), emphasizing complementary cross-modal differences in regions with ambiguous modality preference. Finally, Reliability-Conditioned Reconstruction (RCR) retains trusted observations and injects the recovered cue through a gated residual update. This trust-recover-reconstruct sequence adaptively routes RGB and infrared evidence while preserving the local structure and thermal saliency required for oriented small-vehicle detection.

Experiments on the public DroneVehicle benchmark support the effectiveness of this design. ReDiffNet reaches 85.3\% $\mathrm{mAP}_{50}$, 4.4 points above the highest reported comparison listed in Table~\ref{tab:dronevehicle}. Evaluation on VEDAI provides additional evidence on a second RGB--infrared aerial dataset, while the component-removal study shows measurable contributions from DARL, UGDR, and RCR under a shared training and evaluation protocol.

The contributions of this work are as follows:
\begin{itemize}
  \sloppy
  \item We formulate low-light RGB-IR fusion from the perspective of spatially varying relative modality reliability and introduce reliability-conditioned evidence reconstruction for organizing heterogeneous multimodal observations.
  \item We propose ReDiffNet, which couples DARL, UGDR, and RCR in a trust-recover-reconstruct process that preserves informative observations and recovers complementary cross-modal evidence in regions with ambiguous modality preference.
  \item ReDiffNet achieves 85.3\% $\mathrm{mAP}_{50}$ on DroneVehicle and 73.9\% on VEDAI, while component-wise studies provide supporting evidence for the proposed design.
\end{itemize}
\section{Methods}
\begin{figure*}[t]
  \centering
  \includegraphics[width=\textwidth]{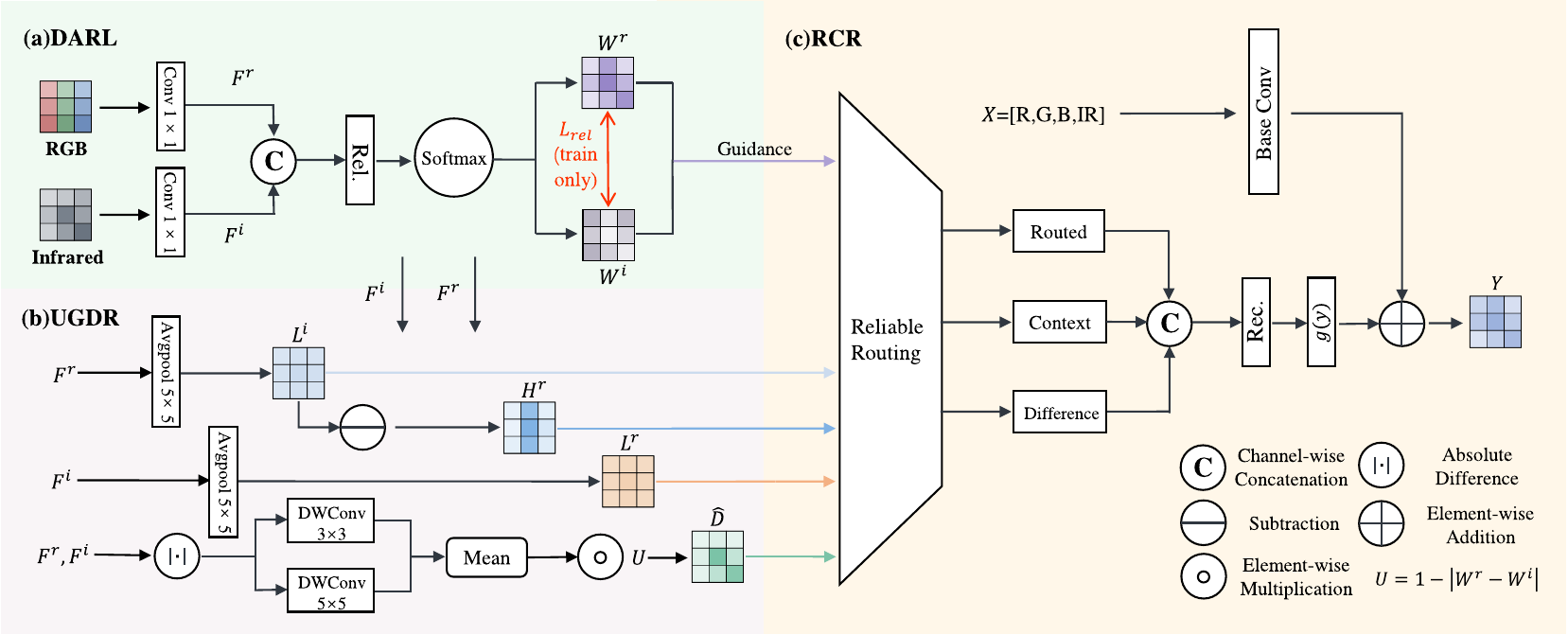}
  \caption{Illustration of the three core modules in ReDiffNet:
(a) DARL for modality reliability estimation,
(b) UGDR for complementary cue recovery, and
(c) RCR for reliable feature reconstruction.}
  \label{fig:architecture}
\end{figure*}

\subsection{Overview}
\begin{sloppypar}
To enhance the reliability and complementarity of multimodal representations under low-light conditions, we propose ReDiffNet, as illustrated in Fig.~\ref{fig:architecture}. ReDiffNet comprises three sequentially coupled components: Degradation-Aware Reliability Learning, Uncertainty-Guided Differential Recovery, and Reliability-Conditioned Reconstruction. Given registered RGB and infrared images, degradation-aware reliability learning estimates the relative local reliability of each modality under spatially varying degradation. Similarity between the two reliability responses defines a reliability-ambiguity cue, which guides differential recovery to strengthen complementary evidence in regions with ambiguous modality preference. Reliability-conditioned reconstruction subsequently preserves trustworthy modality features and employs the recovered differences to compensate for ambiguous observations. Through this progressive process, ReDiffNet produces discriminative representations for low-light UAV-based RGB-infrared oriented small-vehicle detection.
\end{sloppypar}

\subsection{Degradation-Aware Reliability Learning}
Spatially heterogeneous illumination degradation and attenuated thermal responses induce location-dependent variations in modality fidelity. Accordingly, we formulate modality selection as a spatial relative-reliability inference problem and develop a degradation-aware reliability learning component.

\begin{sloppypar}
Given a registered RGB-infrared pair $\{X^r,X^i\}$, modality-specific projections first map the inputs into a shared feature space. The projected features are subsequently aggregated to infer the spatial relative reliability of each modality:
\end{sloppypar}
\begin{equation}
\begin{aligned}
F^m &= \phi_m(X^m), \qquad m \in \{r,i\},\\
[W^r,W^i] &= \operatorname{Softmax}_c\!\left(\phi_{\mathrm{rel}}[F^r\|F^i]\right).
\end{aligned}
\label{eq:darl}
\end{equation}
where $r$ and $i$ denote the RGB and infrared modalities, respectively; $\phi_m(\cdot)$ represents the modality-specific projection; and $\phi_{\mathrm{rel}}(\cdot)$ denotes the reliability estimator. The operator $\|$ indicates channel-wise concatenation, and $\operatorname{Softmax}_c(\cdot)$ normalizes the responses along the modality dimension. Consequently, $W^r(p)+W^i(p)=1$ at each spatial position $p$, and the pair $[W^r(p),W^i(p)]$ represents the normalized relative reliability distribution across the two modalities.

The resulting reliability maps characterize the spatial distribution of relative modality fidelity and provide conditional information for subsequent differential cue recovery. This mechanism prioritizes locally trustworthy observations and improves reliability-aware information selection under low-light conditions.

\subsection{Uncertainty-Guided Differential Recovery}
Small-vehicle boundaries, orientation cues, and thermal responses can be attenuated at different spatial scales. We therefore introduce uncertainty-guided differential cue recovery to enhance complementary information according to the ambiguity of relative modality preference.

Specifically, average pooling is used as a lightweight spatial low-pass approximation to obtain smoothed modality context. Subtracting the smoothed RGB representation from the original RGB feature yields a local-detail residual that emphasizes boundaries and orientation cues. In parallel, the absolute difference between RGB and infrared features is processed by two depthwise convolutions with different receptive fields and then modulated by the ambiguity of the reliability estimates:
\begin{equation}
L^m=\mathcal{P}_5(F^m), \qquad H^r=F^r-L^r, \qquad m\in\{r,i\},
\label{eq:detail}
\end{equation}
\begin{equation}
\begin{aligned}
\widehat{D}={}&\left(1-\left|W^r-W^i\right|\right)\odot\\[-0.3ex]
&\frac{\operatorname{DWConv}_{3}\!\left(\left|F^r-F^i\right|\right)+
\operatorname{DWConv}_{5}\!\left(\left|F^r-F^i\right|\right)}{2}.
\end{aligned}
\label{eq:difference}
\end{equation}
where $\mathcal{P}_5(\cdot)$ denotes $5\times5$ average pooling used as a spatial smoothing operator; $L^m$ is the smoothed contextual representation of modality $m$; and $H^r$ is the RGB local-detail residual. $\operatorname{DWConv}_k(\cdot)$ denotes depthwise convolution with a $k\times k$ kernel, $\odot$ represents element-wise multiplication, and $\widehat{D}$ is the ambiguity-weighted multi-scale differential cue. Here, uncertainty specifically denotes the ambiguity of relative modality preference: similar reliability responses produce a larger ambiguity cue and strengthen differential recovery.

\begin{sloppypar}
This detail-context approximation retains RGB struc\-tural cues and infrared thermal context, while the multi-scale differential branch strengthens complementary evidence in regions where modality preference is ambiguous. Consequently, the component is designed to preserve weak small-vehicle cues while retaining a lightweight spatial formulation.
\end{sloppypar}

\subsection{Reliability-Conditioned Reconstruction}
The final stage converts the estimated reliability and recovered differential cues into a detector-ready representation through reliability-controlled routing followed by a gated residual update.

Reliability-weighted RGB details and infrared thermal responses are first combined with the recovered differential cues. The resulting features are then projected and incorporated into the base representation through a learnable residual gate:
\begin{equation}
R=W^r\odot H^r+W^i\odot L^i+\widehat{D}, \qquad
C=W^r\odot L^r+W^i\odot L^i,
\label{eq:routing}
\end{equation}
\begin{equation}
\begin{aligned}
Z&=\phi_{\mathrm{mix}}[R\|C\|\widehat{D}],\\
Y&=\phi_{\mathrm{base}}[X^r\|X^i]+g(\gamma)\odot\phi_{\mathrm{out}}(Z),\\
g(\gamma)&=0.2\tanh(\gamma).
\end{aligned}
\label{eq:reconstruction}
\end{equation}
where $R$ is the reliability-routed feature, $C$ denotes the modality-weighted contextual feature, and $\phi_{\mathrm{mix}}(\cdot)$ performs feature projection. $\phi_{\mathrm{base}}(\cdot)$ and $\phi_{\mathrm{out}}(\cdot)$ align the base and reconstructed branches, $\|$ denotes channel-wise concatenation, and $\gamma$ is a zero-initialized learnable parameter. The reconstructed representation $Y$ is subsequently supplied to the multi-scale backbone and oriented detection head.

Consequently, trustworthy modality evidence remains dominant, while differential information compensates for ambiguous regions. The gated residual pathway preserves stable information propagation and strengthens low-light small-vehicle representations.

\subsection{Training Loss}
ReDiffNet is trained end-to-end with the oriented detection objective and an auxiliary reliability-supervision loss. The detection loss is computed for both one-to-many and one-to-one assignment branches:
\begin{equation}
\mathcal{L}_{\mathrm{det}}=\sum_{q\in\{\mathrm{o2m},\mathrm{o2o}\}}
\left(\lambda_{\mathrm{box}}\mathcal{L}^{q}_{\mathrm{obb}}+
\lambda_{\mathrm{cls}}\mathcal{L}^{q}_{\mathrm{cls}}\right).
\label{eq:detection-loss}
\end{equation}
\begin{sloppypar}
For synthetically degraded training samples, the mean modality-reliability response is supervised toward the unaffected modality:
\end{sloppypar}
\begin{equation}
\mathcal{L}_{\mathrm{rel}}=-\frac{1}{|\Omega_d|}\sum_{b\in\Omega_d}\log W_b^{y_b},
\qquad
\mathcal{L}_{\mathrm{total}}=\mathcal{L}_{\mathrm{det}}+\lambda_{\mathrm{rel}}\mathcal{L}_{\mathrm{rel}}.
\label{eq:reliability-loss}
\end{equation}
where o2m and o2o denote the one-to-many and one-to-one assignment branches, respectively. $\mathcal{L}_{\mathrm{obb}}$ and $\mathcal{L}_{\mathrm{cls}}$ are the oriented bounding-box regression and classification losses, while $\lambda_{\mathrm{box}}$ and $\lambda_{\mathrm{cls}}$ denote their weights. $\Omega_d$ is the set of degraded samples, $y_b$ is the label of the unaffected modality, and $W_b^{y_b}$ is its spatially averaged reliability response. This supervision encourages DARL to assign higher relative reliability to the unaffected modality, while clean paired inputs are optimized directly by $\mathcal{L}_{\mathrm{det}}$.
\section{Experiments}

\subsection{Experimental Settings}

\noindent\textbf{Datasets.}
We evaluate ReDiffNet on the public DroneVehicle~\cite{r4} and VEDAI~\cite{r20} RGB--infrared benchmarks following their official splits and protocols. DroneVehicle contains registered UAV image pairs annotated for five oriented vehicle classes: car, freight car, truck, bus, and van. The $840\times712$ paired images are cropped by 100 pixels along each border to obtain aligned $640\times512$ inputs, concatenated in the channel order [R,G,B,IR]. Infrared oriented annotations serve as ground truth. The training and validation subsets contain 17,990 and 1,469 pairs, respectively. Main benchmark results are reported on the official test set.

VEDAI provides paired visible and infrared aerial images from suburban and rural scenes. Results are reported on its official test set following the prescribed protocol.

\noindent\textbf{Evaluation Metrics.}
Following previous DroneVehicle methods~\cite{r4,r6,r7}, we use oriented
$\mathrm{mAP}_{50}$ as the primary evaluation metric, where AP is
computed at an IoU threshold of 0.5 and averaged over all object
categories.
For controlled ablation experiments, we additionally report Precision,
Recall, $F_1$, and $\mathrm{mAP}_{50:95}$, where
$\mathrm{mAP}_{50:95}$ averages AP over IoU thresholds from 0.50 to
0.95 with a step of 0.05.

\noindent\textbf{Implementation Details.}
ReDiffNet is implemented in PyTorch and integrated into an Ultralytics-based oriented detection framework. The reconstructed representation is supplied to the shared multi-scale backbone and oriented detection head, and all reported variants use the same detection scaffold. Training is conducted on a single NVIDIA GeForce RTX 4090 GPU for 30 epochs at $640\times640$ resolution with a batch size of 24. We use AdamW with an initial learning rate of 0.001111, weight decay of $5\times10^{-4}$, a three-epoch warm-up, cosine learning-rate decay, and random seed 0. Data augmentation comprises translation, scale, and horizontal flipping to preserve paired-image correspondence.

For reliability supervision, a training pair uses the original paired observations with probability 0.70, while RGB or infrared degradation is sampled with probability 0.15 each. The auxiliary reliability loss has weight $\lambda_{\mathrm{rel}}=0.05$. Main benchmark evaluation uses the original test images with standard single-pass inference.

\subsection{Main Results}

\par\medskip
\noindent\begin{minipage}{\columnwidth}
\makeatletter\def\@captype{table}\makeatother
\centering
\fontsize{9}{9.8}\selectfont
\setlength{\tabcolsep}{1pt}
\begin{tabular*}{\linewidth}{@{\extracolsep{\fill}}llccc@{}}
\toprule
Method & Pub. + Year & RGB & Infrared & $\mathrm{mAP}_{50}$\\
\midrule
RetinaNet~\cite{lin2017focal} & ICCV 2017 & \cmark & \xmark & 47.1\\
R3Det~\cite{yang2021r3det} & AAAI 2021 & \cmark & \xmark & 60.8\\
S2ANet~\cite{han2021align} & TGRS 2021 & \cmark & \xmark & 61.0\\
Faster R-CNN~\cite{ren2016faster} & TPAMI 2017 & \cmark & \xmark & 55.9\\
RoITransformer~\cite{ding2019learning} & CVPR 2019 & \cmark & \xmark & 61.6\\
Oriented R-CNN~\cite{xie2021oriented} & ICCV 2021 & \cmark & \xmark & 60.8\\
YOLO26s~\cite{r13} & arXiv 2026 & \cmark & \xmark & 74.5\\
\midrule
RetinaNet~\cite{lin2017focal} & ICCV 2017 & \xmark & \cmark & 54.5\\
R3Det~\cite{yang2021r3det} & AAAI 2021 & \xmark & \cmark & 62.3\\
S2ANet~\cite{han2021align} & TGRS 2021 & \xmark & \cmark & 67.5\\
Faster R-CNN~\cite{ren2016faster} & TPAMI 2017 & \xmark & \cmark & 64.2\\
RoITransformer~\cite{ding2019learning} & CVPR 2019 & \xmark & \cmark & 65.5\\
Oriented R-CNN~\cite{xie2021oriented} & ICCV 2021 & \xmark & \cmark & 67.0\\
\midrule
CIAN~\cite{r14} & IF 2019 & \cmark & \cmark & 70.8\\
MBNet~\cite{r15} & ECCV 2020 & \cmark & \cmark & 71.9\\
TSFADet~\cite{r5} & ECCV 2022 & \cmark & \cmark & 73.9\\

SLBAF-Net~\cite{r16} & MTA 2023 & \cmark & \cmark & 76.1\\
C$^2$Former~\cite{r6} & TGRS 2024 & \cmark & \cmark & 74.2\\
DMM~\cite{r7} & TGRS 2025 & \cmark & \cmark & 77.2\\
CoDAF~\cite{r18} & ASOC 2026 & \cmark & \cmark & 78.6\\
IGIANet~\cite{r19} & AAAI 2026 & \cmark & \cmark & 80.9\\
\midrule
ReDiffNet (Ours) &  & \cmark & \cmark & \textbf{85.3}\\
\bottomrule
\end{tabular*}

\caption{Comparison with state-of-the-art methods on DroneVehicle.
RGB and Infrared indicate the input modalities used by each method.
All $\mathrm{mAP}_{50}$ values are percentages.}
\label{tab:dronevehicle}
\end{minipage}
\par\medskip

\noindent\textbf{Results on DroneVehicle.}
As shown in Table~\ref{tab:dronevehicle}, ReDiffNet achieves
85.3\% $\mathrm{mAP}_{50}$ on DroneVehicle.
Compared with the best-performing RGB-only and infrared-only detectors,
ReDiffNet improves $\mathrm{mAP}_{50}$ by 10.8 and 17.8 percentage
points, respectively.
Among the compared methods, YOLO26s~\cite{r13} achieves 74.5\%
$\mathrm{mAP}_{50}$, while DMM~\cite{r7}, CoDAF~\cite{r18},
and IGIANet~\cite{r19} achieve 77.2\%, 78.6\%, and 80.9\%,
respectively.
ReDiffNet reaches 85.3\%, exceeding the highest reported comparison
in Table~\ref{tab:dronevehicle} by 4.4 percentage points.

\par\medskip
\noindent\begin{minipage}{\columnwidth}
\makeatletter\def\@captype{table}\makeatother
\centering
\fontsize{9}{9.8}\selectfont
\setlength{\tabcolsep}{1pt}
\begin{tabular*}{\linewidth}{@{\extracolsep{\fill}}lccc@{}}
\toprule
Method & RGB & Infrared & $\mathrm{mAP}_{50}$\\
\midrule
RetinaNet~\cite{lin2017focal} & \cmark & \xmark & 20.7\\
S2ANet~\cite{han2021align} & \cmark & \xmark & 44.5\\
Faster R-CNN~\cite{ren2016faster} & \cmark & \xmark & 61.5\\
RoITransformer~\cite{ding2019learning} & \cmark & \xmark & 65.4\\
Oriented R-CNN~\cite{xie2021oriented} & \cmark & \xmark & 66.4\\
\midrule
RetinaNet~\cite{lin2017focal} & \xmark & \cmark & 18.7\\
S2ANet~\cite{han2021align} & \xmark & \cmark & 40.0\\
Faster R-CNN~\cite{ren2016faster} & \xmark & \cmark & 55.4\\
RoITransformer~\cite{ding2019learning} & \xmark & \cmark & 60.5\\
Oriented R-CNN~\cite{xie2021oriented} & \xmark & \cmark & 60.9\\
\midrule
C$^2$Former+S2ANet~\cite{r6} & \cmark & \cmark & 55.6\\
CMAFF+Oriented R-CNN~\cite{r21} & \cmark & \cmark & 69.2\\
\midrule
ReDiffNet (Ours) & \cmark & \cmark & \textbf{73.9}\\
\bottomrule
\end{tabular*}
\caption{Comparison on the VEDAI dataset.
All values are $\mathrm{mAP}_{50}$ .}
\label{tab:vedai}
\end{minipage}
\par\medskip

\noindent\textbf{Results on VEDAI.}
As shown in Table~\ref{tab:vedai}, ReDiffNet achieves
73.9\% $\mathrm{mAP}_{50}$ on VEDAI.
Compared with the best-performing RGB-only and infrared-only detectors,
ReDiffNet improves $\mathrm{mAP}_{50}$ by 7.5 and 13.0 percentage
points, respectively.
Compared with multimodal methods, ReDiffNet exceeds
C$^2$Former+S2ANet~\cite{r6} by 18.3 points and
CMAFF+Oriented R-CNN~\cite{r21} by 4.7 points.
These results provide additional evidence for the proposed method on a
second RGB--infrared aerial detection benchmark.

\par\medskip
\noindent\begin{minipage}{\columnwidth}
\makeatletter\def\@captype{table}\makeatother
\centering
\fontsize{9}{9.8}\selectfont
\setlength{\tabcolsep}{1pt}
\begin{tabular*}{\linewidth}
{@{\extracolsep{\fill}}lcccc@{}}
\toprule
Method & Params (M) & FLOPs (G) & FPS& $\mathrm{mAP}_{50}$ \\
\midrule
TSFADet~\cite{r5}
& 104.7 & 109.8 & 18.6 &73.9 \\

CAGTDet~\cite{r22}
& -- & 120.6 & 17.8 & 74.6
\\

C$^2$Former~\cite{r6}
& 100.8 & 89.9 & -- &74.2\\

DMM+S$^2$A-Net~\cite{r7}
& 87.97 & -- & -- & 79.4\\

CoDAF~\cite{r18}
& 67.3 & 224.9 & 58.1& 78.6\\
\midrule
ReDiffNet (Ours)
& \textbf{10.53} & 27.42 & 106.9 & \textbf{85.3} \\
\bottomrule
\end{tabular*}
\caption{Reported model size and detection performance on DroneVehicle.
All $\mathrm{mAP}_{50}$ values are percentages.}
\label{tab:complexity}
\end{minipage}
\par\medskip

\par\medskip
\noindent\begin{minipage}{\columnwidth}
\makeatletter\def\@captype{table}\makeatother
\centering
\fontsize{9}{9.8}\selectfont
\setlength{\tabcolsep}{1pt}
\begin{tabular*}{\linewidth}{@{\extracolsep{\fill}}lccccc@{}}
\toprule
Method & Precision & Recall & $F_1$
& $\mathrm{mAP}_{50}$ & $\mathrm{mAP}_{50:95}$\\
\midrule
DARL removed
& 80.18 & 77.95 & 79.05 & 81.71 & 67.30\\
UGDR removed
& 79.53 & 75.95 & 77.70 & 80.48 & 66.16\\
RCR removed
& 79.12 & 76.86 & 77.98 & 80.91 & 66.62\\
Detection objective only
& 79.12 & 78.50 & 78.80 & 83.10 & 68.40\\
\midrule
Full model
& \textbf{81.79}
& \textbf{80.84}
& \textbf{81.31}
& \textbf{85.30}
& \textbf{70.48}\\
\bottomrule
\end{tabular*}
\caption{Ablation study on DroneVehicle.
All variants use the same training and evaluation protocol.}
\label{tab:ablation}
\end{minipage}
\par\medskip
\noindent\textbf{Model Complexity.} As shown in Table~\ref{tab:complexity}, ReDiffNet contains 10.53M parameters, about 15.6\% of CoDAF's~\cite{r18} (67.3M) and one-tenth of TSFADet's~\cite{r5} (104.7M). ReDiffNet achieves 85.3\% $\mathrm{mAP}_{50}$ with this compact architecture, 5.9 percentage points above the highest reported result listed in the table. This comparison highlights the favorable balance between detection performance and overall model size.

\subsection{Ablation Study}

\noindent\textbf{Component Ablation.}
As shown in Table~\ref{tab:ablation}, all three components contribute to the overall performance of ReDiffNet. The DARL-removed variant achieves 81.71\% $\mathrm{mAP}_{50}$ and 67.30\% $\mathrm{mAP}_{50:95}$, compared with 85.30\% and 70.48\% for the full model. This difference reflects the role of spatial reliability estimation in organizing modality-dependent features.

The UGDR-removed variant achieves 80.48\% $\mathrm{mAP}_{50}$ and 66.16\% $\mathrm{mAP}_{50:95}$, showing the largest decrease among the three component variants. This result highlights the importance of ambiguity-guided differential recovery for modeling complementary cross-modal cues.

The RCR-removed variant obtains 80.91\% $\mathrm{mAP}_{50}$ and 66.62\% $\mathrm{mAP}_{50:95}$, indicating that reliability-conditioned reconstruction further improves multimodal feature integration.

Training with the detection objective alone yields 83.10\% $\mathrm{mAP}_{50}$, whereas the full training objective reaches 85.30\%. The 2.20-point improvement shows the benefit of auxiliary reliability supervision. Overall, DARL, UGDR, and RCR provide complementary functions within ReDiffNet.

\par\medskip
\noindent\begin{minipage}{\columnwidth}
\makeatletter\def\@captype{table}\makeatother
\centering
\fontsize{9}{9.8}\selectfont
\setlength{\tabcolsep}{1pt}
\begin{tabular*}{\linewidth}{@{\extracolsep{\fill}}llccc@{}}
\toprule
Condition & Factor & $\mathrm{mAP}_{50}$ & $\mathrm{mAP}_{50:95}$ & Drop$_{50}$\\
\midrule
Clean & -- & 85.21 & 70.19 & --\\
\midrule
RGB darkening & 0.75 & 84.80 & 69.80 & 0.41\\
RGB darkening & 0.50 & 84.00 & 68.89 & 1.21\\
RGB darkening & 0.30 & 82.77 & 67.63 & 2.44\\
\midrule
IR contrast reduction & 0.80 & 85.10 & 70.05 & 0.12\\
IR contrast reduction & 0.60 & 84.66 & 69.62 & 0.55\\
IR contrast reduction & 0.40 & 82.78 & 67.80 & 2.43\\
\midrule
RGB darkening (local) & 0.50 & 85.15 & 70.04 & 0.07\\
IR contrast (local) & 0.60 & 85.06 & 70.01 & 0.15\\
\bottomrule
\end{tabular*}

\caption{ReDiffNet under synthetic modality degradation on the DroneVehicle validation set; Drop$_{50}$ is in percentage points.}
\label{tab:degradation}
\end{minipage}
\par\medskip

\noindent\textbf{Controlled Degradation Analysis.}
We evaluate the same ReDiffNet checkpoint on the DroneVehicle validation split under synthetic single-modality perturbations. As shown in Table~\ref{tab:degradation}, performance decreases gradually as RGB darkening or infrared contrast reduction becomes stronger. At the strongest evaluated settings, $\mathrm{mAP}_{50}$ decreases by 2.44 and 2.43 percentage points for RGB darkening and infrared contrast reduction, respectively. Under local perturbations, ReDiffNet retains 85.15\% and 85.06\% $\mathrm{mAP}_{50}$, with drops of only 0.07 and 0.15 points. Clean denotes the original inputs; the factors scale RGB intensity or infrared contrast, while the local variants affect the central third along each spatial dimension.

\section{Conclusion}

In this paper, we propose ReDiffNet, a reliability-conditioned differential representation network for low-light UAV-based RGB--infrared oriented vehicle detection. ReDiffNet follows a trust-recover-reconstruct process that combines relative reliability learning, ambiguity-guided differential recovery, and reliability-conditioned reconstruction. Experiments on DroneVehicle and VEDAI support the effectiveness of the proposed framework. Future work will explore its extension to broader multimodal perception tasks.

\clearpage
\section*{Acknowledgment}

This work was supported by the Liaoning Revitalization Talents Program (XLYCE2504014) and the Scientific Research Innovation Capability Support Project for Young Faculty (3132025232). The authors gratefully acknowledge the support
provided for the implementation and experimental evaluation of this work.

The authors report an absence of relevant financial or non-financial
interests to disclose. ChatGPT (OpenAI) was used solely for language
translation, polishing, and grammar enhancement during manuscript
preparation.

\bibliographystyle{IEEEbib}
\bibliography{references}

\end{document}